\documentclass[10pt,twocolumn,letterpaper]{article}

\usepackage[pagenumbers]{wacv} 

\usepackage{multirow}
\usepackage[table]{xcolor}
\usepackage{algorithm}
\usepackage{algpseudocode}
\usepackage{tabularx}
\usepackage{array}
\definecolor{wacvblue}{rgb}{0.21,0.49,0.74}
\usepackage[breaklinks,colorlinks,allcolors=wacvblue]{hyperref}

\def\wacvPaperID{2080} 
\def\confName{WACV}
\def\confYear{2027}

\title{Semi-Dense Matching Uncertainty Is Not Just Local Confidence}

\author{
Khoa Hoang$^{1,2}$ \quad
Hoang-Tuan Nguyen$^{3}$ \quad
Huong Ninh$^{1,2}$ \quad
Hai Tran$^{2}$ \quad
Long Q. Tran$^{1}$\\[0.5em]
$^{1}$University of Engineering and Technology, Vietnam National University, Hanoi\\
$^{2}$Optoelectronics Center, Viettel Aerospace Institute, Viettel Group\\
$^{3}$National Yang Ming Chiao Tung University, Taiwan
}

\begin{document}
\maketitle
\begin{abstract}
Reliable semi-dense matching is essential for modern geometric vision systems. Designed under a coarse-to-fine paradigm, it achieves an optimal balance between performance and computational cost. However, existing methods often struggle to provide well-quantified uncertainties, where catastrophic coarse-assignment failures are ignored, leading to truncated error distributions and severely misjudged geometric estimations. In this paper, we propose a lightweight, post-hoc overall uncertainty estimation framework that introduces a two-component calibrated Laplace mixture  model with only 9 learnable parameters. The objective is to explicitly capture both the sharp local refinement noise and the broader tail of coarse-assignment failures. We introduce the Coarse-success posterior Refit (CoRe) method, a geometric refitting module that utilizes the posterior probability of coarse-assignment success as soft correspondence weights. Extensive experiments show that our method consistently improves downstream geometric accuracy across various pretrained matchers and robust estimators with minimal computational overhead. Our code is available at
\href{https://github.com/khoavpt/Probabilistic-matching}
{\texttt{https://github.com/khoavpt/Probabilistic\\-matching}}.
\end{abstract}

\section{Introduction}
\label{sec:intro}

\begin{figure}[t]
  \centering
   \includegraphics[width=1.0\linewidth]{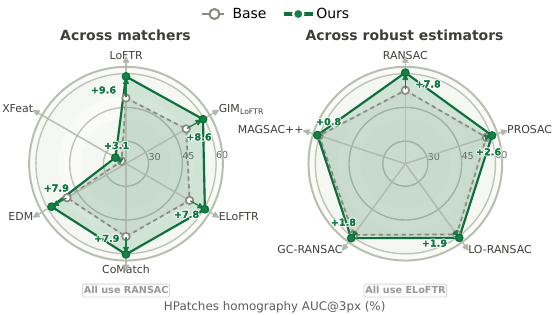}
    \caption{\textbf{Coarse-success posterior refit improves downstream homography estimation.}
    Radar plots compare HPatches AUC@3px before and after our refit. The left panel varies the matcher with RANSAC fixed, while the right varies the robust estimator with ELoFTR fixed.}
   \label{fig:mixture}
\end{figure}

Image matching is a key foundation of modern geometric vision systems, critical for downstream applications such as Structure-from-Motion (SfM) \cite{schonberger2016structure}, visual localization \cite{sarlin2019coarse, sattler2012image}, and Simultaneous Localization and Mapping (SLAM) \cite{mur2015orb, mur2017orb}. Recently, semi-dense feature matching methods \cite{sun2021loftr, wang2024eloftr, DBLP:conf/iccv/LiRP25} have demonstrated remarkable success by adopting a coarse-to-fine paradigm. These architectures balance efficiency and performance by restricting global correspondence search to coarse features, requiring the fine stage to predict only local offsets within a local correspondence window \cite{sun2021loftr, wang2024eloftr, DBLP:conf/iccv/LiRP25,  DBLP:conf/iccv/LiLTZM25}. 

However, coarse-to-fine matchers frequently struggle to provide well-quantified uncertainties for their predictions. This two-stage design fundamentally fractures the error distribution into two distinct sources: local refinement errors and coarse assignment failures. Existing methods only model the former, either implicitly through the fine-stage heatmaps ~\cite{sun2021loftr, DBLP:conf/cvpr/PotjeC0MN24, wang2024eloftr, DBLP:conf/iccv/LiRP25}, or, more recently, explicitly through predicted local uncertainty ~\cite{DBLP:journals/corr/abs-2603-04869}. Typically, treating uncertainty as a local patch confidence score implicitly assumes absolute certainty in the initial coarse assignment. However, in heavily occluded, textureless, or repetitive regions, coarse hypotheses frequently fail, placing the true correspondence entirely outside the local refinement window. Thus, strictly local uncertainty measurement inherently ignores the risk of coarse-level catastrophic failures, yielding an overall truncated error distribution. For downstream robust geometric estimation, this under-qualified uncertainty is a critical vulnerability, as an overconfident, severely misplaced match can bias an estimated model far more than a highly uncertain but correctly localized one.

Our key observation is therefore simple: \textbf{matching uncertainty is not just local confidence}. We argue that a useful correspondence likelihood must model both the sharp local inlier mode and the broader tail of coarse assignment failures. To achieve this, in this paper, we propose a very lightweight, post-hoc matching uncertainty estimation method for semi-dense matchers. Rather than relying on just local confidence, we view their matching process under a probabilistic lens and approximate the overall matching error as a two-component Laplace mixture. A calibrated gate dynamically combines these two components using uncertainty cues already produced by the pretrained matchers. Our method requires no retraining of the base matcher, and only 9 calibration parameters are fitted, making our method a simple plug-and-play module for existing pipelines. 

To leverage this comprehensive uncertainty model, we introduce CoRe (Coarse-success posterior Refit), which uses the posterior probability of coarse-assignment success as a soft correspondence weight. Instead of discarding correspondences via hard inlier thresholds after an initial robust estimation (e.g., RANSAC \cite{DBLP:journals/cacm/FischlerB81}), CoRe performs a single soft-weighted refit, allowing the final geometric model to penalize catastrophic coarse failures while still retaining valid correspondences, with only modest runtime overhead. As shown in Fig.~\ref{fig:mixture}, CoRe consistently improves downstream geometry across different matchers and robust estimators, demonstrating the practical value of modeling the high-error tails caused by coarse-assignment failures that are often ignored.

Our contributions are summarized as follows:
(i) We introduce an overall matching uncertainty estimation as a two-component Laplace calibrated mixture model, capturing both local refinement noise and catastrophic coarse failures.
(ii) The proposed framework requires only 9 learnable calibration parameters for dynamically gating uncertainty cues, serving as a plug-and-play module for off-the-shelf semi-dense matching.
(iii) We propose CoRe (Coarse-success posterior Refit), which leverages the calibrated error model for robust geometry estimation. 
(iv) Extensive experiments demonstrate the effectiveness of our proposed method, achieving significant performance improvements.

\section{Related Work}
\label{sec:related-works}
\begin{figure}[t]
  \centering
    \includegraphics[width=1.0\linewidth]{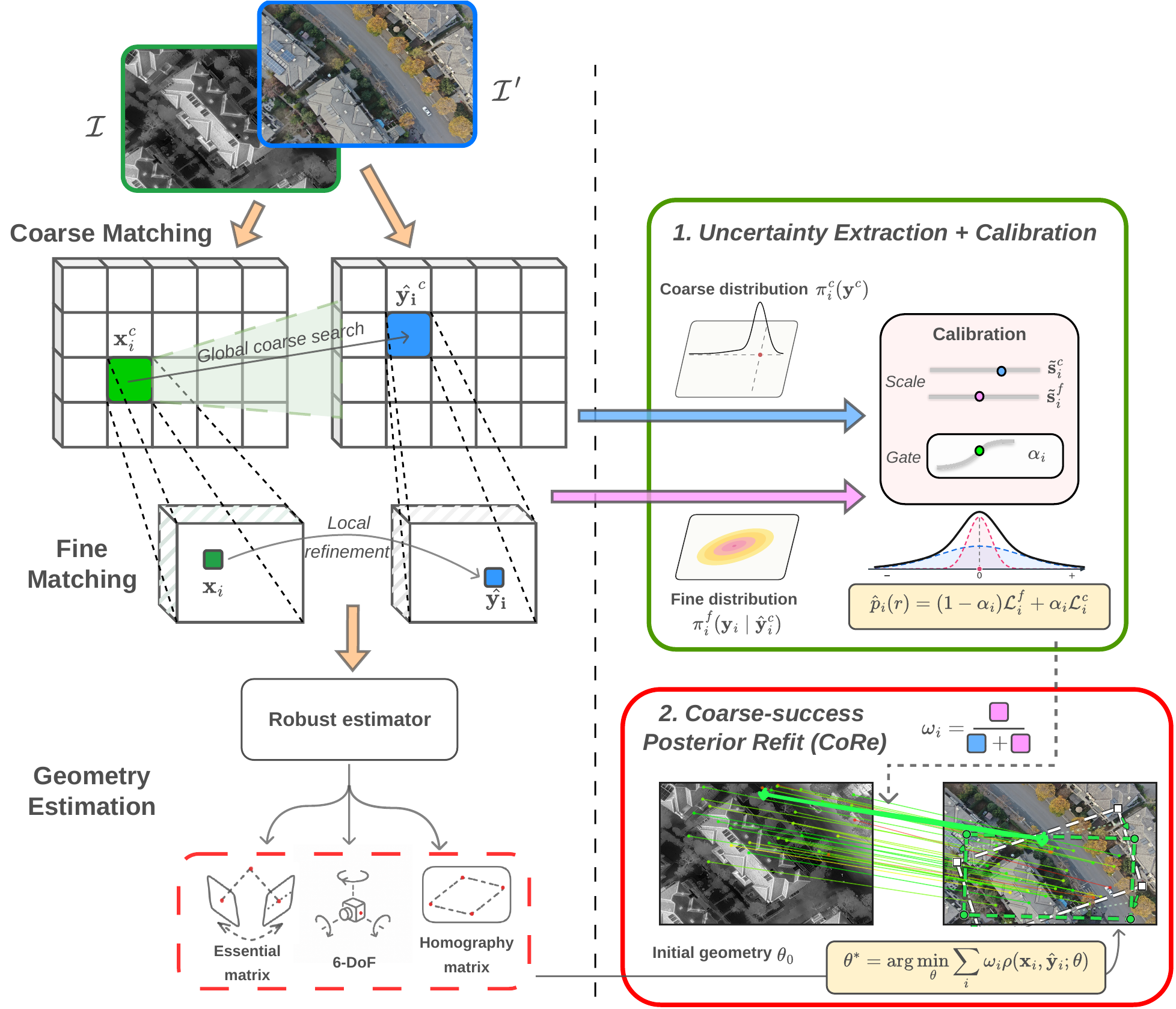}   
    \caption{
    \textbf{Overview of the proposed framework.}
    Left: the conventional coarse-to-fine matching and robust geometry estimation pipeline. Right: our framework calibrates coarse- and fine-level cues into a two-component error likelihood (Alg.~\ref{alg:calibration}), which is used to estimate coarse-assignment success posteriors and perform our posterior-weighted geometric refit (CoRe; Alg.~\ref{alg:weighted_refit}).
    }
  \label{fig:robust_estimator}
  \vspace{-10pt}
\end{figure}

\paragraph{Coarse-to-fine Matching and Uncertainty Modeling.}
Semi-dense matching strikes an optimal balance between computational efficiency and matching performance, which not only fully utilizes the entire image space but also avoids overly dense pixel-level calculations by adopting the coarse-to-fine manner. 
LoFTR \cite{sun2021loftr} and its variants \cite{wang2022matchformer, chen2022aspanformer, tang2022quadtree, giang2023topicfm, DBLP:conf/iccv/LiRP25, DBLP:conf/iccv/LiLTZM25, DBLP:journals/corr/abs-2603-04869} apply the Transformer \cite{vaswani2017attention} to enhance local features and then refine its correspondence in a local window. 
However, their matchers derive predictions and uncertainty strictly from fine-stage windows. They implicitly assume absolute certainty in the initial coarse assignment, frequently failing to properly evaluate overall predictive uncertainty. While rigorous uncertainty estimation has driven significant progress in other computer vision tasks \cite{mac2012learning, kondermann2007adaptive, kondermann2008statistical, gal2016dropout}, in image matching, it has mostly been explored by dense architectures. For instance, PDC-Net \cite{truong2021learning} and its extension \cite{truong2023pdc} utilize a constrained mixture model to predict dense correspondences along with certainty estimates. DKM~\cite{edstedt2023dkm} formulates match certainty as the likelihood of geometric consistency, trained directly under 3D supervision. RoMAv2~\cite{edstedt2025romav2} predicts the Gaussian covariance of 2D matching residuals via Cholesky factorization.  Motivated by this, we propose a framework that adapts probabilistic mixture modeling to explicitly capture both the local refinement error and the catastrophic coarse failures for overall uncertainty estimation in semi-dense pipelines.

\paragraph{Robust geometric estimation.}

Robust estimators have long been an essential requirement for estimating 3D geometry from noisy correspondences~\cite{hartley2003multiple}. While RANSAC~\cite{DBLP:journals/cacm/FischlerB81} is the foundational algorithm, it relies heavily on a brittle, hard inlier-outlier threshold. Advanced variants have systematically improved this process: PROSAC~\cite{DBLP:conf/cvpr/ChumM05} accelerates convergence by sampling highly confident points first, GC-RANSAC~\cite{DBLP:conf/cvpr/BarathM18} incorporates spatial coherence among points, MAGSAC~\cite{DBLP:conf/cvpr/BarathMN19} and MAGSAC++~\cite{DBLP:conf/cvpr/BarathNIM20} mitigate the hard threshold issue by scoring points across a continuous sliding scale of noise levels. Other studies also demonstrate the effectiveness of incorporating explicit scoring functions into the geometric estimation pipeline \cite{shekhovtsov2025ransac, edstedt2025romav2}. Building upon these principles, our method bridges probabilistic calibration and robust geometry by introducing a posterior-weighted refit. We incorporate the posterior probability of coarse-assignment success as weights for the geometric refit process, utilizing residual information of all correspondences that is mostly discarded by standard RANSAC pipelines.

\paragraph{Calibration and probabilistic correspondences.}
Modern neural networks often produce poorly calibrated, overconfident
predictions~\cite{3305381.3305518}. In classification, simple post-hoc methods such as temperature scaling can substantially improve calibration without retraining the model~\cite{3305381.3305518}. For continuous regression, calibration is commonly defined through the coverage of predictive quantiles or credible intervals~\cite{DBLP:conf/icml/KuleshovFE18}. However, quantile
calibration is a marginal criterion and does not necessarily ensure that the predicted conditional distribution matches the true target distribution for individual inputs. Distribution calibration addresses this stronger requirement by aiming to align the predicted distribution to the empirical conditional distribution of the target variables~\cite{DBLP:conf/icml/SongDKF19}. Motivated by these calibration principles, we transform uncertainty cues extracted from the matching process into metrically calibrated spatial error distributions, without requiring network retraining.

\section{Methodology}
\label{sec:method}

\providecommand{\bestcell}[1]{\cellcolor{green!18}#1} \providecommand{\secondcell}[1]{\cellcolor{yellow!25}#1} \providecommand{\thirdcell}[1]{\cellcolor{orange!18}#1} \newcommand{\aucgain}[2]{#1\,{\scriptsize$\rightarrow$}\,#2} \newcommand{\locauc}[6]{#1/#2/#3\,{\scriptsize$\rightarrow$}\,#4/#5/#6} \providecommand{\gain}[1]{\textbf{#1}}

\begin{figure*}[t]
  \centering
\includegraphics[width=0.97\linewidth]{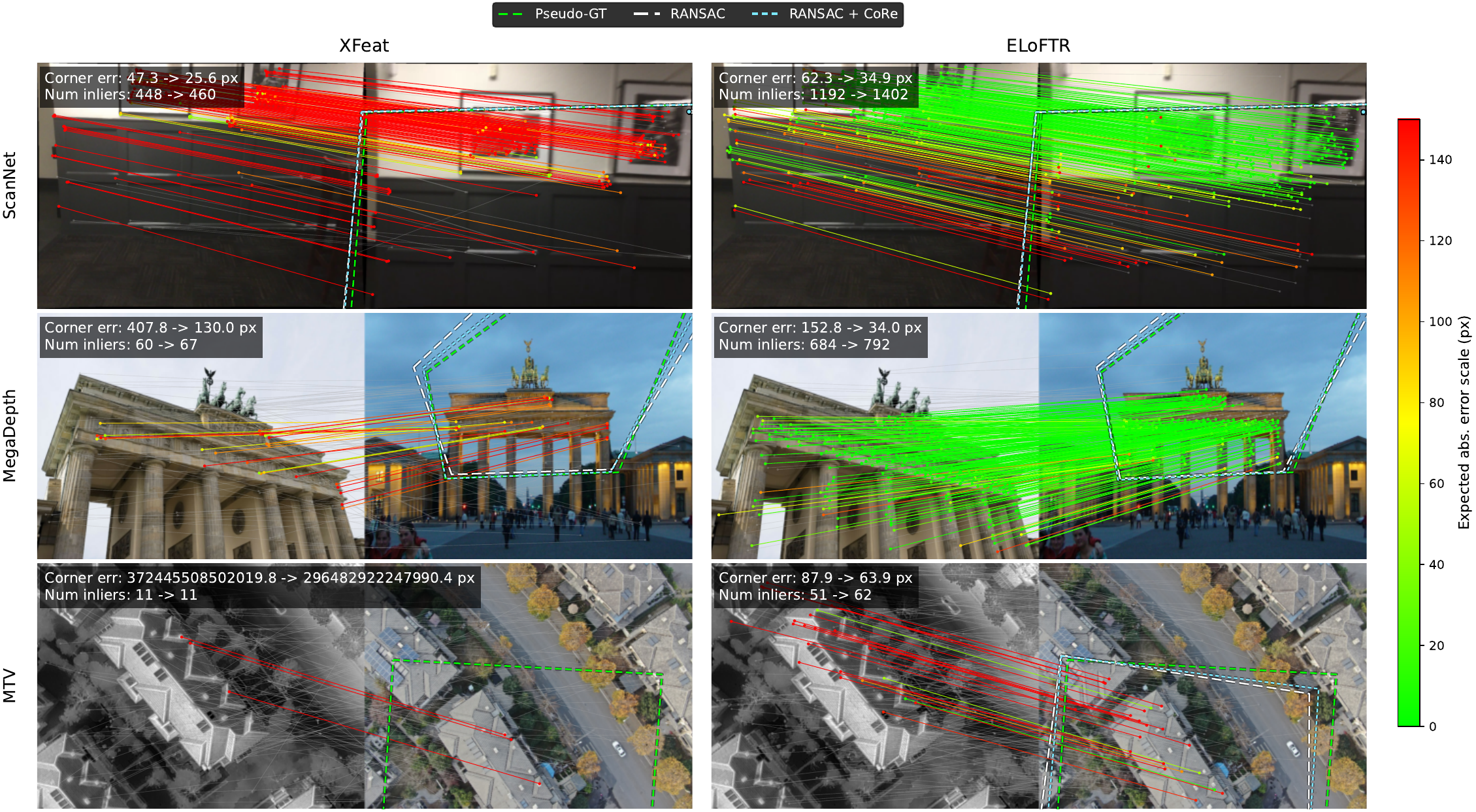}
  \caption{
    \textbf{Qualitative homography refit.}
    This figure shows the effect of applying CoRe to homography estimation. Rows show increasingly difficult pairs from top to bottom, and columns compare stronger matchers from left to right. Matches are colored by Expected abs. error scale (px), from low uncertainty in green to high uncertainty in red, with gray matches de-emphasized. CoRe is applied on all matches including gray outliers. Details are recorded in the top-left of each image pair.
  }
  \label{fig:qualiative}
  \vspace{-5pt}
\end{figure*}

\subsection{Problem Statement}
\label{subsec:prem}

Following the coarse-to-fine paradigm, given an image pair $(I,I')$, correspondences are first established on a low-resolution feature grid and subsequently refined within a local window around the predicted coarse match~\cite{sun2021loftr}. This design is computationally efficient, as global correspondence search is performed only at the coarse level, while the refinement stage is restricted to a small local region.

Let $\mathbf{x}_i\in\mathbb{R}^2$ be a point in the first image $I$, and $\mathbf{y}_i\in\mathbb{R}^2$ denote its ground-truth correspondence in the second image ${I'}$. To keep notation light, we write all distributions with subscript $i$ as conditioned on the source point $\mathbf{x}_i$. From a probabilistic perspective, the coarse stage predicts a distribution $\pi_i^c(\mathbf{y}^c)$ over candidate coarse cells $\mathbf{y}^c$ in the second image $I'$. This distribution is typically obtained from similarities between low-resolution features~\cite{sun2021loftr,wang2024eloftr}. The coarse correspondence is then selected as
\begin{equation}
\hat{\mathbf{y}}^c_i
=
\arg\max_{\mathbf{y}^c}
\pi_i^c(\mathbf{y}^c).
\end{equation}

The overall matching uncertainty combines the variance of the coarse prior and the conditional variance of the refinement step.
Typically, conditioned on an initial coarse hypothesis $\hat{\mathbf{y}}_i^c$, the refinement stage estimates a local probability distribution. However, this only captures the conditional distribution $\pi_i^f(\mathbf{y}_i \mid \hat{\mathbf{y}}_i^c)$, strictly characterizing the uncertainty in the local displacement.

Relying exclusively on $\pi_i^f(\mathbf{y}_i \mid \hat{\mathbf{y}}_i^c)$ yields a truncated marginal distribution, implicitly treating the selected coarse assignment as correct with probability one. If the underlying coarse distribution is highly ambiguous or assigned incorrectly, the true correspondence may lie completely outside the local support of the conditional density. This coarse-assignment error is fundamentally missed by purely local estimators \cite{sun2021loftr, wang2024eloftr, DBLP:conf/iccv/LiRP25, DBLP:conf/iccv/LiLTZM25}. A rigorous probabilistic framework must estimate the full marginal distribution over the entire spatial domain, capturing both coarse-level ambiguity and local refinement inaccuracies.

\subsection{Correspondence Error Modeling}
\label{subsec:uncertain}

\begin{algorithm}[t]
\caption{Post-hoc error model calibration}
\label{alg:calibration}
\begin{algorithmic}[1]
\Require calibration set \(\mathcal D_{\mathrm{cal}}\), frozen matcher \(M\)
\State initialize \(\Theta=\{\mathbf a,\mathbf b,\mathbf w,\beta\}\)
\Repeat
    \ForAll{\((I,I')\in\mathcal D_{\mathrm{cal}}\)}
        \State \(\{(\mathbf x_i,\hat{\mathbf y}_i,\tilde{\mathbf s}_i^f,
        \tilde{\mathbf s}_i^c,\boldsymbol\phi_i)\}_i \leftarrow M(I,I')\)
        \State \(\mathbf r_i \leftarrow \hat{\mathbf y}_i-\mathbf y_i\)
        \State \(\mathbf s_i^f \leftarrow
        \sqrt{\mathbf b}\odot\tilde{\mathbf s}_i^f,\quad
        \mathbf s_i^c \leftarrow
        \sqrt{\mathbf a}\odot\tilde{\mathbf s}_i^c\)
        \State \(\alpha_i \leftarrow
        \sigma(\mathbf w^\top\boldsymbol\phi_i+\beta)\)
    \EndFor
    \State update \(\Theta\) to minimize
    \(-\sum_i\log \hat p_i(\mathbf r_i;\Theta)\)
\Until{convergence}
\State \Return \(\Theta\)
\end{algorithmic}
\end{algorithm}

\begin{algorithm}[t]
\caption{CoRe: Coarse-success posterior geometric Refit}
\label{alg:weighted_refit}
\begin{algorithmic}[1]
\Require image pair $(I,I')$, frozen matcher $M$, calibrated $\Theta = \{\mathbf{a}, \mathbf{b}, \mathbf{w}, \beta\}$, robust estimator $R$
\State \(\{(\mathbf x_i,\hat{\mathbf y}_i,\tilde{\mathbf s}_i^f,
\tilde{\mathbf s}_i^c,\boldsymbol\phi_i)\}_i \leftarrow M(I,I')\)
\State $\theta_0 \leftarrow R(\{(\mathbf{x}_i,\hat{\mathbf{y}}_i)\}_i)$
\ForAll{$i$}
    \State \(\mathbf s_i^f \leftarrow
    \sqrt{\mathbf b}\odot\tilde{\mathbf s}_i^f,\quad
    \mathbf s_i^c \leftarrow
    \sqrt{\mathbf a}\odot\tilde{\mathbf s}_i^c\)
    \State $\alpha_i \leftarrow \sigma(\mathbf{w}^{\top}\boldsymbol{\phi}_i+\beta)$
    \State $\mathbf{r}_i^{(0)}\leftarrow \hat{\mathbf{y}}_i-\Pi_{\theta_0}(\mathbf{x}_i)$
    \State $\omega_i \leftarrow \frac{(1-\alpha_i) \text{Lap}(\mathbf{r}_i^{(0)}|\mathbf{0}, \mathbf{s}_{i}^{f})}{(1-\alpha_i) \text{Lap}(\mathbf{r}_i^{(0)}|\mathbf{0}, \mathbf{s}_{i}^{f}) + \alpha_i \text{Lap}(\mathbf{r}_i^{(0)}|\mathbf{0}, \mathbf{s}_{i}^{c})}$
\EndFor
\State $\theta^\star \leftarrow \arg\min_{\theta}\sum_i \omega_i\rho(\mathbf{x}_i,\hat{\mathbf{y}}_i;\theta)$
\State \Return $\theta^\star$
\end{algorithmic}
\end{algorithm}

Reliably modeling the error distribution of predicted correspondences is crucial for high-precision geometric estimation and provides a foundation for future uncertainty estimation of the resulting geometric model~\cite{hartley2003multiple}.

\paragraph{Matching Errors Decomposition.}
To address the limitations discussed above, we start by viewing the final correspondence error as being induced by the coarse-to-fine matching process. Let $\mathbf{r}_i=\hat{\mathbf{y}}_i-\mathbf{y}_i$ denote the signed residual of correspondence $i$, where $\hat{\mathbf{y}}_i$ is the predicted point and $\mathbf{y}_i$ is the ground-truth correspondence in the second image $I'$. The residual distribution can be expressed by marginalizing over possible coarse assignments:
\begin{equation}
p_i(\mathbf{r})
=
\sum_{\mathbf{y}^c}
\pi_i^c(\mathbf{y}^c)\,
p_i(\mathbf{r}\mid \mathbf{y}^c).
\end{equation}



The overall correspondence error depends critically on the quality of the coarse assignment.
Assuming an accurate coarse cell prediction (i.e., $\mathbf{y}^c = \mathbf{y}^{c,*}_i$), the remaining error corresponds strictly to the local refinement residual. In this scenario, the refinement module provides an estimate of the conditional residual distribution, denoted as $p_i(\mathbf{r}\mid \mathbf{y}^{c,*}_i) = p_i^f(\mathbf{r}\mid \mathbf{y}^{c,*}_i)$. In contrast, when the coarse prediction is incorrect, the overall error is dominated by the coarse-assignment failure. Given that the spatial envelope of the coarse error strictly dominates the spatial support of the refinement, we can practically approximate the residual distribution as $p_i(\mathbf{r} \mid \mathbf{y}^c \neq \mathbf{y}^{c,*}_i) \approx p_i^c(\mathbf{r})$. 
By weighting these two mutually exclusive states following the coarse accuracy probability $\pi_i^c(\mathbf{y}^{c,*}_i)$, we formulate the residual distribution as a two-component mixture model:
\begin{equation}
p_i(\mathbf{r})
\approx
\pi_i^c(\mathbf{y}^{c,*}_i)\,
p_i^f(\mathbf{r}\mid \mathbf{y}^{c,*}_i)
+
\bigl(1-\pi_i^c(\mathbf{y}^{c,*}_i)\bigr)\,
p_i^c(\mathbf{r}).
\label{eq:two_component_error}
\end{equation}

This approximation enables practical integration with existing coarse-to-fine matchers \cite{sun2021loftr, wang2024eloftr, DBLP:conf/iccv/LiRP25, DBLP:conf/iccv/LiLTZM25}. Since these matchers only run the fine refinement module for the selected coarse hypothesis, modeling every conditional term $p(\mathbf{r}_i\mid \mathbf{y}^c)$ is not feasible. Our proposed mixture model still keeps the fine component for the selected coarse match while using a separate coarse-error component to account for failures outside the local refinement window.

\paragraph{Error Model Calibration.}

\begin{figure}[t]
  \centering
    \includegraphics[width=0.96\linewidth]{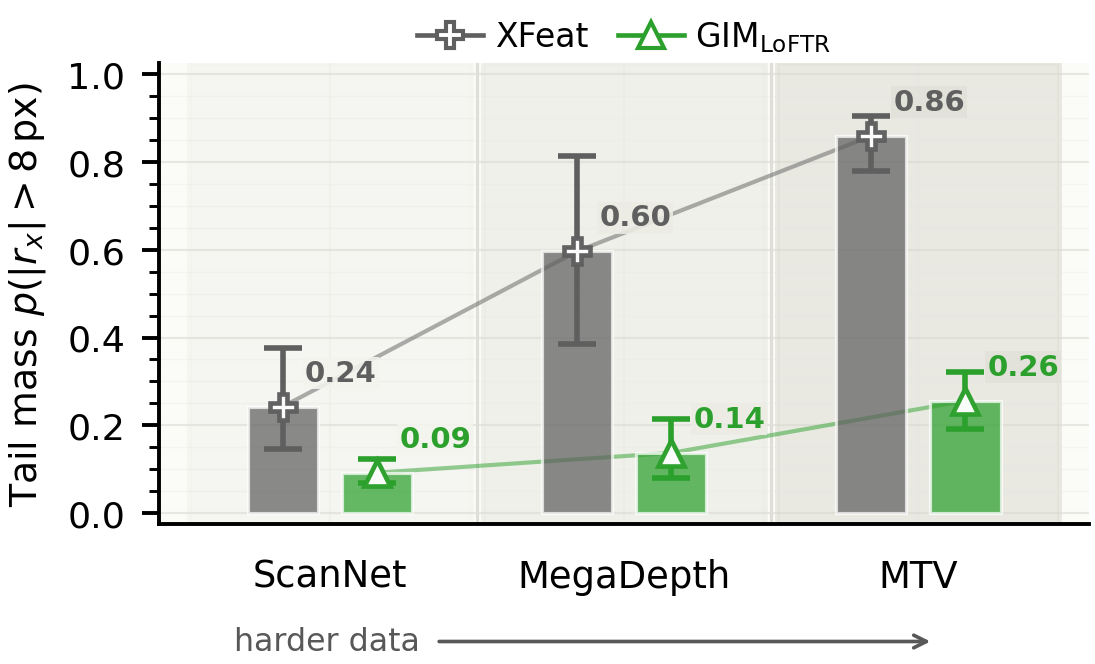}
    \caption{\textbf{Tail probability assignment.} The model dynamically assigns higher tail mass ($P(|r_x| > 8\text{px})$) on harder datasets and for weaker matchers, demonstrating adaptive calibration.}
  \label{fig:tail_behavior}
  \vspace{-20pt}
\end{figure}

For simplicity, we model both of these components using Laplace distributions. To compensate for modeling approximations, we jointly calibrate the scales of each component and the gating variable between them. Under this assumption, specifically for each match $i$, we define:
\begin{equation}
\hat p_i(\mathbf{r})
=
(1-\alpha_i)\,
\mathrm{Lap}(\mathbf{r}\mid \mathbf{0},\mathbf{s}^f_i)
+
\alpha_i\,
\mathrm{Lap}(\mathbf{r}\mid \mathbf{0},\mathbf{s}^c_i),
\label{eq:laplace_mixture}
\end{equation}

Here, the calibrated fine and coarse scales, denoted as $\mathbf{s}^f_i, \mathbf{s}^c_i \in \mathbb{R}^2_{+}$, are derived from their raw estimates via:
\begin{equation}
\mathbf{s}^f_i
=
\sqrt{\mathbf{b}}\odot \tilde{\mathbf{s}}^f_i,
\qquad
\mathbf{s}^c_i
=
\sqrt{\mathbf{a}}\odot \tilde{\mathbf{s}}^c_i,
\label{eq:scale_calibration}
\end{equation}
where $\mathbf{a}, \mathbf{b} \in \mathbb{R}^2_{+}$ are learnable calibration parameters. These calibration parameters compensate for the fact that heatmap variance, predicted sigma, and coarse distribution spread are not directly expressed as calibrated pixel-error scales. Here, $\tilde{\mathbf{s}}^f_i$ is the raw fine scale extracted from the fine refinement output (e.g., the standard deviation of a local heatmap \cite{sun2021loftr, DBLP:conf/iccv/LiRP25}), and $\tilde{\mathbf{s}}^c_i$ is the raw coarse scale extracted from the coarse matching distribution. Crucially, $\tilde{\mathbf{s}}^f_i$ and $\tilde{\mathbf{s}}^c_i$ are directly extracted from pre-trained semi-dense image matching models, without any fine-tuning. 

In addition, we define $\alpha_i$ as the calibrated mixture weight of the coarse-error component.  Intuitively, $\alpha_i$ should be small when the coarse assignment is reliable, but increase when the coarse distribution is ambiguous or the match confidence is low. We therefore model it as a sigmoid over a linear combination of coarse-level cues:
\begin{equation}
\alpha_i
=
\sigma\!\left(
\mathbf w^\top\boldsymbol\phi_i+\beta
\right),
\label{eq:gate}
\end{equation}
where \(\boldsymbol{\phi}_i\) comprises the image-size-normalized coarse scale and the normalized peak value of the coarse matching distribution. Both \(\mathbf{w}\) and \(\beta\) are learnable calibration parameters. 

Following the principle of distribution-level calibration~\cite{DBLP:conf/icml/SongDKF19}, we fit the calibration parameters $\Theta = \{\mathbf{a}, \mathbf{b}, \mathbf{w}, \beta\}$ by minimizing the negative log-likelihood, which is a strictly proper scoring rule~\cite{Gneiting01032007}.
\begin{equation}
\mathcal{L}_{\mathrm{calib}}
=
-\sum_i
\log \hat p_i(\mathbf{r}_i).
\label{eq:nll}
\end{equation}
Only the calibration parameters are optimized, while the base matcher remains frozen. This keeps the method very lightweight and allows the same calibration procedure to be applied to existing coarse-to-fine matchers. 

\begin{table*}[t]
\centering
\caption{\textbf{Homography estimation on HPatches and MTV.} Comparison of original RANSAC results and with CoRe-refined results (\emph{Base $\rightarrow$ w/ CoRe}). Parameters are calibrated on MegaDepth and transferred zero-shot.}
\label{tab:homography_hpatches_mtv}
\vspace{-4pt}
\scriptsize
\setlength{\tabcolsep}{3.0pt}
\renewcommand{\arraystretch}{1.12}
\begin{tabular*}{0.99\textwidth}{@{\extracolsep{\fill}}lcccc@{\hspace{1.1em}}cccc@{}}
\toprule
\multirow{2}{*}{\textbf{Matcher}} &
\multicolumn{4}{c}{\textbf{HPatches} $\uparrow$} &
\multicolumn{4}{c}{\textbf{MTV} $\uparrow$} \\
\cmidrule(lr){2-5} \cmidrule(lr){6-9}
& \textbf{@1px} & \textbf{@3px} & \textbf{@5px} & \textbf{@10px}
& \textbf{@10px} & \textbf{@30px} & \textbf{@50px} & \textbf{@100px} \\
\midrule
LoFTR \cite{sun2021loftr}
& \aucgain{21.37}{\textbf{28.80}}
& \aucgain{49.17}{\textbf{58.72}}
& \aucgain{61.40}{\textbf{70.19}}
& \aucgain{75.18}{\textbf{81.54}}
& \aucgain{1.08}{\textbf{1.83}}
& \aucgain{4.77}{\textbf{6.40}}
& \aucgain{7.05}{\textbf{8.65}}
& \aucgain{10.01}{\textbf{11.65}} \\

GIM\textsubscript{LoFTR} \cite{xuelun2024gim}
& \aucgain{24.82}{\textbf{29.87}}
& \aucgain{50.87}{\textbf{59.47}}
& \aucgain{62.23}{\textbf{71.11}}
& \aucgain{76.53}{\textbf{82.87}}
& \aucgain{0.44}{\textbf{1.31}}
& \aucgain{3.63}{\textbf{6.27}}
& \aucgain{6.43}{\textbf{9.85}}
& \aucgain{11.41}{\textbf{15.24}} \\

EfficientLoFTR \cite{wang2024eloftr}
& \aucgain{29.14}{\textbf{32.68}}
& \aucgain{52.59}{\textbf{60.43}}
& \aucgain{64.17}{\textbf{71.24}}
& \aucgain{77.30}{\textbf{82.45}}
& \aucgain{1.10}{\textbf{1.78}}
& \aucgain{4.67}{\textbf{6.11}}
& \aucgain{7.04}{\textbf{8.63}}
& \aucgain{10.43}{\textbf{11.93}} \\

CoMatch \cite{DBLP:conf/iccv/LiLTZM25}
& \aucgain{27.47}{\textbf{32.40}}
& \aucgain{52.35}{\textbf{60.26}}
& \aucgain{64.00}{\textbf{71.27}}
& \aucgain{77.22}{\textbf{82.67}}
& \aucgain{0.96}{\textbf{1.65}}
& \aucgain{4.31}{\textbf{5.47}}
& \aucgain{6.54}{\textbf{7.63}}
& \aucgain{9.44}{\textbf{10.43}} \\

EDM \cite{DBLP:conf/iccv/LiRP25}
& \aucgain{24.39}{\textbf{29.45}}
& \aucgain{50.25}{\textbf{58.15}}
& \aucgain{62.07}{\textbf{69.80}}
& \aucgain{76.01}{\textbf{81.56}}
& \aucgain{1.32}{\textbf{2.23}}
& \aucgain{5.87}{\textbf{7.49}}
& \aucgain{8.82}{\textbf{10.80}}
& \aucgain{13.10}{\textbf{15.17}} \\

XFeat* \cite{DBLP:conf/cvpr/PotjeC0MN24}
& \aucgain{0.23}{\textbf{0.33}}
& \aucgain{22.30}{\textbf{25.43}}
& \aucgain{40.19}{\textbf{47.38}}
& \aucgain{62.18}{\textbf{69.51}}
& \aucgain{0.46}{\textbf{0.91}}
& \aucgain{2.39}{\textbf{3.77}}
& \aucgain{3.67}{\textbf{5.39}}
& \aucgain{5.83}{\textbf{7.86}} \\
\bottomrule
\end{tabular*}
\vspace{-10pt}
\end{table*}

\subsection{CoRe: Coarse-Success Posterior Geometric Refit}


Standard robust estimation pipelines typically perform a final refinement step by re-estimating the geometric model using the set of identified inliers (i.e., binary hard-assignment) \cite{DBLP:journals/cacm/FischlerB81, hartley2003multiple}. This approach treats all inliers equally and discards all residual information of the outlier correspondences. To overcome this rigid inlier/outlier truncation, we propose a probabilistic, uncertainty-aware weighting scheme evaluated over all matches. By exploiting our calibrated error model, this method turns residual information from all correspondences into soft reliability weights, enabling a more accurate geometric refit.

To this end, we introduce CoRe, a geometric refinement step that uses the coarse-success posterior from our calibrated error model as a soft weight for each correspondence. Let $\theta_0$ denote an initial geometric model estimated using a robust off-the-shelf estimation method. We compute the reprojection residual of each correspondence as follows:
\begin{equation}
\mathbf{r}_i^{(0)}
=
\hat{\mathbf{y}}_i
-
\Pi_{\theta_0}(\mathbf{x}_i),
\label{eq:geom_residual}
\end{equation}

where $\Pi_{\theta_0}(\mathbf{x}_i)$ denotes the projection of $\mathbf{x}_i$ induced by the model $\theta_0$. The fine-stage refinement is strictly conditioned on the accuracy of the coarse hypothesis. Therefore, a match is geometrically reliable only when the initial coarse assignment is correct. Recalling the formulation in Eq.~\eqref{eq:two_component_error}, we identify these reliable correspondences by computing the posterior probability that the initial coarse assignment succeeded ($\hat{\mathbf{y}}^c_i = \mathbf{y}^{c,*}_i$), given the residual under the initial estimate $\mathbf{r}_i^{(0)}$. Applying Bayes' theorem, the correspondence weight $\omega_i$ is derived as:
\begin{align}
\omega_i 
&= \text{Pr}(\hat{\mathbf{y}}^c_i = \mathbf{y}^{c,*}_i \mid \mathbf{r}_i^{(0)}) \nonumber \\
&= \frac{\pi_i^c(\mathbf{y}^{c,*}_i)\, p_i^f(\mathbf{r}_i^{(0)}\mid \mathbf{y}^{c,*}_i)}{\pi_i^c(\mathbf{y}^{c,*}_i)\, p_i^f(\mathbf{r}_i^{(0)}\mid \mathbf{y}^{c,*}_i) + \bigl(1-\pi_i^c(\mathbf{y}^{c,*}_i)\bigr)\, p_i^c(\mathbf{r}_i^{(0)})} \label{eq:weight_exact} \\
&\approx \frac{(1-\alpha_{i}) \mathcal{L}_{i}^{f}}{(1-\alpha_{i}) \mathcal{L}_{i}^{f} + \alpha_{i} \mathcal{L}_{i}^{c}}. \label{eq:weight_approx}
\end{align}

Since the true underlying residual distributions are unknown, we substitute them in Eq.~\eqref{eq:weight_approx} with our empirical Laplace mixture model. We approximate the fine and coarse distributions with our calibrated Laplace components, $\mathcal{L}_{i}^{f} = \text{Lap}(\mathbf{r}_i^{(0)}|\mathbf{0}, \mathbf{s}_{i}^{f})$ and $\mathcal{L}_{i}^{c} = \text{Lap}(\mathbf{r}_i^{(0)}|\mathbf{0}, \mathbf{s}_{i}^{c})$, and model the prior probability of a coarse failure, $1 - \pi_i^c(\mathbf{y}^{c,*}_i)$, using our calibrated gating variable $\alpha_i$. The final geometric model is then obtained by a weighted refit over all matches:
\begin{equation}
\theta^{*} = \arg\min_{\theta} \sum_{i} \omega_{i} \rho(\mathbf{x}_{i}, \hat{\mathbf{y}}_{i}; \theta),
\end{equation}
where $\rho$ is the geometric fitting loss (e.g., squared reprojection error). By applying these posterior weights globally, CoRe allows potential coarse failures to be softly downweighted rather than discarded completely.

Notably, CoRe serves as a plug-and-play module that preserves the original sampling, scoring, and stopping criteria of off-the-shelf robust estimators \cite{DBLP:journals/cacm/FischlerB81, DBLP:conf/dagm/ChumMK03, DBLP:conf/cvpr/ChumM05, DBLP:conf/cvpr/BarathM18, DBLP:conf/cvpr/BarathNIM20} without any modifications. Since it only needs a single uncertainty extraction and one final weighted refit as additional computation, CoRe achieves these geometric improvements with minimal runtime overhead, as shown in Fig.~\ref{fig:runtime_bar}

\section{Experiments}
\label{sec:exp}

\providecommand{\bestcell}[1]{\cellcolor{green!18}#1}
\providecommand{\secondcell}[1]{\cellcolor{yellow!25}#1}

\providecommand{\gain}[1]{\textbf{#1}}

\begin{table}[t]
\centering
\caption{\textbf{Visual localization on Aachen v1.1.} Percentage of localized queries within pose-error thresholds. CoRe improves accuracy in most settings.}
\label{tab:visual_localization}
\vspace{-4pt}
\scriptsize
\setlength{\tabcolsep}{3.0pt}
\renewcommand{\arraystretch}{1.12}
\begin{tabular*}{0.97\linewidth}{@{\extracolsep{\fill}}lcc@{}}
\toprule
\multirow{2}{*}{\textbf{Matcher}} &
\textbf{Day} $\uparrow$ &
\textbf{Night} $\uparrow$ \\
\cmidrule(lr){2-3}
&
\multicolumn{2}{c}{\textbf{$(0.25\mathrm{m},2^\circ)$ / $(0.5\mathrm{m},5^\circ)$ / $(5.0\mathrm{m},10^\circ)$}} \\
\midrule
LoFTR
& \locauc{85.3}{91.3}{94.5}{\gain{86.9}}{\gain{92.0}}{\gain{95.1}}
& \locauc{65.4}{83.2}{91.6}{\gain{69.1}}{\gain{86.9}}{\gain{92.1}} \\

GIM\textsubscript{LoFTR}
& \locauc{86.9}{92.8}{97.0}{\gain{87.9}}{\gain{93.4}}{\gain{97.1}}
& \locauc{68.6}{80.1}{89.0}{67.0}{\gain{81.2}}{\gain{90.1}} \\

EfficientLoFTR
& \locauc{85.4}{91.3}{94.9}{\gain{86.7}}{\gain{92.6}}{\gain{95.0}}
& \locauc{73.3}{84.3}{92.7}{72.8}{\gain{85.3}}{92.7} \\

CoMatch
& \locauc{85.2}{91.4}{95.1}{85.1}{91.1}{\gain{95.5}}
& \locauc{66.5}{85.9}{94.2}{\gain{70.2}}{\gain{87.4}}{94.2} \\

EDM
& \locauc{87.3}{93.1}{96.8}{\gain{88.0}}{\gain{93.2}}{\gain{97.0}}
& \locauc{70.2}{89.5}{97.9}{\gain{73.3}}{\gain{90.6}}{97.9} \\

XFeat
& \locauc{63.2}{70.3}{77.8}{\gain{67.1}}{\gain{74.0}}{\gain{79.6}}
& \locauc{23.6}{34.0}{45.5}{\gain{28.8}}{\gain{38.2}}{\gain{49.7}} \\
\bottomrule
\end{tabular*}
\vspace{-10pt}
\end{table}

We evaluate the practical utility of the proposed calibrated model beyond simply reporting an uncertainty value. The experiments address two core questions. First, does the mixture model accurately fit the empirical error distribution, capturing both fine and coarse matching failures? Second, does the model provide meaningful geometric information such that refitting the geometric model toward correspondences with a high posterior probability of coarse-assignment success improves downstream estimation?

\subsection{Implementation Details}
\label{subsec:settings}

Unless otherwise stated, each matcher is calibrated post-hoc on 200 held-out MegaDepth~\cite{MDLi18} pairs at $1280{\times}960$, with the matcher frozen and only the calibration parameters optimized using Powell's method~\cite{powell1964efficient}. The parameters are transferred to all downstream datasets without finetuning.

All comparisons use the same matcher outputs and the same robust estimation settings for the baseline and CoRe. CoRe preserves the estimator's sampling, scoring, and stopping
criteria~\cite{DBLP:journals/cacm/FischlerB81,DBLP:conf/dagm/ChumMK03,
DBLP:conf/cvpr/ChumM05,DBLP:conf/cvpr/BarathM18,DBLP:conf/cvpr/BarathNIM20}, adding only one posterior-weighted refit. In practice, we use an axis-factorized implementation with separate one-dimensional Laplace mixtures for the horizontal and vertical residuals (see Supp. Sec.~\ref{sec:axis-implementation}). The final CoRe weight is therefore computed as the product of the two axis-wise coarse-success posteriors.

\begin{figure*}[t]
  \centering
  \includegraphics[width=0.96\linewidth]{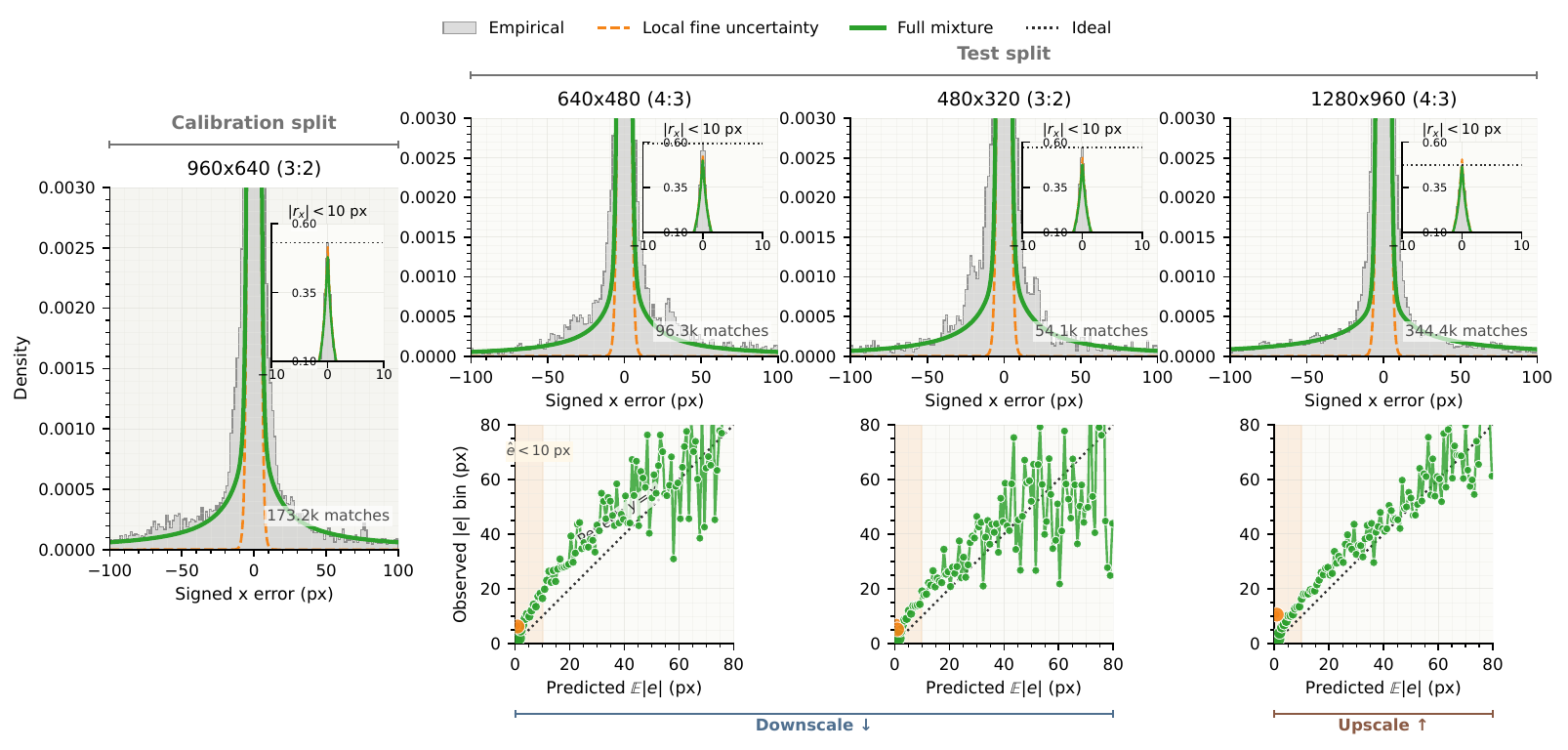}
    \caption{
    \textbf{Resolution transfer of the calibrated error model.}
    Calibrated on 200 held-out MegaDepth pairs at $960{\times}640$ and evaluated
    without retuning on MegaDepth-1500 at other resolutions.
    Top: signed residual distribution for one coordinate axis, with fine-only and
    mixture densities; insets show the central peak.
    Bottom: reliability plots of predicted $\mathbb{E}[e]$ versus observed $e=\frac{1}{2}(|r_x|+|r_y|)$; the dotted line denotes perfect calibration. Unlike fine-only uncertainty, which is bounded by the refinement window, our mixture tracks observed errors across resolutions.
    }
  \label{fig:pdf_transfer}
  \vspace{-10pt}
\end{figure*}

\subsection{Zero-Shot Evaluation on Downstream Tasks}
\label{subsec:downstream}

We evaluate the calibrated likelihood on two downstream settings that require accurate geometric estimation from image matches: planar homography estimation with HPatches~\cite{DBLP:conf/cvpr/BalntasLVM17}, MTV~\cite{DBLP:journals/remotesensing/LiuLYCZPZ23} and visual localization with Aachen v1.1~\cite{DBLP:conf/cvpr/SattlerMTTHSSOP18}. We apply CoRe to the pipelines of several recent semi-dense matchers, including LoFTR~\cite{sun2021loftr}, GIM\textsubscript{LoFTR}~\cite{xuelun2024gim}, EfficientLoFTR~\cite{wang2024eloftr}, CoMatch~\cite{DBLP:conf/iccv/LiLTZM25}, EDM~\cite{DBLP:conf/iccv/LiRP25}, and XFeat~\cite{DBLP:conf/cvpr/PotjeC0MN24}.

\paragraph{Homography Estimation.}

We test homography estimation on the widely used HPatches dataset~\cite{DBLP:conf/cvpr/BalntasLVM17} and MTV~\cite{DBLP:journals/remotesensing/LiuLYCZPZ23}, a highly challenging out-of-domain dataset pairing aerial thermal and RGB images. HPatches is evaluated at a $1280{\times}960$ resolution, while MTV is evaluated at $960{\times}640$ to account for its imprecise ground-truth annotations.

We use the algebraic DLT objective for geometric fitting, resulting in a weighted DLT refit. We collect the mean reprojection error of corner points and report the AUC at 1, 3, 5, and 10 px thresholds for HPatches, and 10, 30, 50 and 100 px thresholds for MTV.

As shown in Tab.~\ref{tab:homography_hpatches_mtv}, CoRe consistently improves homography estimation across all matchers and error thresholds. By actively utilizing residual information from all correspondences, including assumed outliers, CoRe provides a more optimal fit that safely down-weights coarse failures. This drives pronounced gains at broader thresholds while maintaining strong precision at 1px. Furthermore, consistent improvements on the highly ambiguous cross-modal MTV dataset validate the robustness of our zero-shot calibration under severe domain shift.

\begin{table}[t]
\centering
\vspace{-4pt}
\scriptsize
\renewcommand{\arraystretch}{1.12}

\begin{tabular*}{\linewidth}{@{\extracolsep{\fill}}lccccc}
\toprule
\multirow{2}{*}{\textbf{Refit weight}} &
\multicolumn{5}{c}{\textbf{Homography est. AUC} $\uparrow$} \\
\cmidrule(lr){2-6}
& \textbf{@1px} & \textbf{@3px} & \textbf{@5px}
& \textbf{@10px} & \textbf{@30px} \\
\midrule

Original RANSAC \cite{DBLP:journals/cacm/FischlerB81}
    & 29.14 & 52.59 & 64.17 & 77.30 & 89.27 \\
Uniform refit
    & 19.36 & 43.85 & 54.46 & 66.19 & 78.86 \\
Binary inliers/outliers
    & \thirdcell{33.11} & 56.87 & 67.50 & 79.55 & 90.20 \\

\addlinespace[2pt]
\rowcolor{gray!8}
\multicolumn{6}{l}{\hspace{2pt}\textit{Residual-based continuous weighting}} \\
Huber refit
    & 32.80 & \thirdcell{59.41} & \thirdcell{69.93}
    & \thirdcell{81.33} & 90.72 \\
Laplace--Laplace refit
    & \bestcell{34.66} & 59.20 & 69.56
    & 81.06 & \thirdcell{90.89} \\

\addlinespace[2pt]
\rowcolor{gray!8}
\multicolumn{6}{l}{\hspace{2pt}\textit{Matcher-derived uncertainty weighting}} \\
Coarse confidence refit
    & 25.07 & 49.04 & 59.22 & 70.17 & 81.35 \\
Raw fine std
    & \secondcell{33.65} & \secondcell{59.47} & \secondcell{70.06}
    & \secondcell{81.60} & \secondcell{91.08} \\
\textbf{CoRe}
    & 32.85 & \bestcell{60.50} & \bestcell{71.25}
    & \bestcell{82.44} & \bestcell{91.32} \\

\bottomrule
\end{tabular*}
\caption{\textbf{Different refit weighting strategies comparison on HPatches.}
Comparison of refit weighting strategies using EfficientLoFTR. CoRe's posterior-based weighting achieves the highest AUC from $3$px onward.}
\label{tab:ablation_refit}
\end{table}

\paragraph{Visual Localization.}
We evaluate 6-DoF camera pose estimation on the Aachen Day-Night v1.1 benchmark.
Images are resized to \(1180{\times}896\) for evaluation. We use a hierarchical pipeline inspired by HLoc~\cite{sarlin2019coarse}: each query is
matched against the top-50 NetVLAD~\cite{DBLP:conf/cvpr/ArandjelovicGTP16}
retrieved references to form 2D--3D correspondences, and the initial pose is estimated using PnP inside RANSAC \cite{DBLP:journals/cacm/FischlerB81}. 

To apply CoRe, we evaluate the calibrated mixture on the initial reprojection residuals and perform one weighted nonlinear refinement minimizing squared reprojection error over all raw matches. We report the percentage of queries satisfying the standard Aachen pose-error thresholds.

As shown in Tab.~\ref{tab:visual_localization}, while the performance increases are more modest than in the homography experiments, CoRe still improves localization accuracy in most settings. These gains are most evident in the Night track.

\subsection{Ablation Studies}
\label{subsec:ablation}

This section validates the modeling and algorithmic choices of our method. For all ablations, we use EfficientLoFTR~\cite{wang2024eloftr} as the frozen base matcher and follow the calibration protocol in Sec.~\ref{subsec:settings}.

\paragraph{Refit Weighting Strategy.}
Tab.~\ref{tab:ablation_refit} ablates refit weighting on HPatches~\cite{DBLP:conf/cvpr/BalntasLVM17}, separating the benefit of continuous robust weighting from that of matcher-derived uncertainty. We compare CoRe with one-step Huber M-estimation~\cite{Huber1992,Holland01011977} (thresholded at $2$px) and a residual-only Laplace--Laplace mixture whose scales and prior are estimated from the initial RANSAC residuals and inlier ratio. Both outperform uniform and binary refits, confirming the benefit of continuous weighting. CoRe achieves the best AUC from $3$px onward, outperforming both residual-only and raw matcher-uncertainty weighting, demonstrating the additional benefit of calibrated coarse-aware reliability.

The slight AUC@$1$ drop mainly occurs for already accurate initial estimates, where refitting offers little room for improvement and small perturbations can cross the strict $1$px threshold. This matches our sensitivity analysis (Supp.~Sec.~B.1), showing negligible gains below $1$px but clear gains for larger errors.

\paragraph{Robustness to Base Estimators.}
To ensure CoRe is orthogonal to the initial robust solver, we apply it to the outputs of LO-RANSAC~\cite{DBLP:conf/dagm/ChumMK03}, PROSAC~\cite{DBLP:conf/cvpr/ChumM05}, GC-RANSAC~\cite{DBLP:conf/cvpr/BarathM18}, and MAGSAC++~\cite{DBLP:conf/cvpr/BarathNIM20} (Tab.~\ref{tab:homography_estimators}). CoRe consistently improves
AUC@$3$px--$10$px across all estimators. Mirroring our previous findings, the marginal drop at 1px occurs primarily because applying a global weighted refit across all matches introduces slight spatial smoothing that can perturb these near-perfect fits.

\begin{table}[t]
\centering
\caption{\textbf{Homography estimation across robust estimators.}
We apply CoRe to various RANSAC variants using EfficientLoFTR matches.
CoRe provides consistent gains across solvers, with small trade-offs at the strict 1px threshold.}
\label{tab:homography_estimators}
\vspace{-4pt}
\scriptsize
\setlength{\tabcolsep}{1.5pt}
\renewcommand{\arraystretch}{1.08}
\providecommand{\aucg}[2]{#1{\scriptsize$\!\rightarrow\!$}#2}
\begin{tabular*}{\linewidth}{@{\extracolsep{\fill}}lcccc}
\toprule
\multirow{2}{*}{\textbf{Estimator}} &
\multicolumn{4}{c}{\textbf{Homography AUC} $\uparrow$} \\
\cmidrule(lr){2-5}
& \textbf{1px} & \textbf{3px} & \textbf{5px} & \textbf{10px} \\
\midrule
RANSAC \cite{DBLP:journals/cacm/FischlerB81}
& \aucg{29.14}{\textbf{32.68}}
& \aucg{52.59}{\textbf{60.43}}
& \aucg{64.17}{\textbf{71.24}}
& \aucg{77.30}{\textbf{82.45}} \\

LO-RANSAC \cite{DBLP:conf/dagm/ChumMK03}
& \aucg{\textbf{33.89}}{32.73}
& \aucg{58.82}{\textbf{60.69}}
& \aucg{69.85}{\textbf{71.56}}
& \aucg{81.30}{\textbf{82.77}} \\

PROSAC \cite{DBLP:conf/cvpr/ChumM05}
& \aucg{\textbf{33.76}}{32.78}
& \aucg{57.82}{\textbf{60.46}}
& \aucg{67.98}{\textbf{71.11}}
& \aucg{79.41}{\textbf{82.09}} \\

GC-RANSAC \cite{DBLP:conf/cvpr/BarathM18}
& \aucg{\textbf{33.99}}{32.85}
& \aucg{59.08}{\textbf{60.86}}
& \aucg{69.97}{\textbf{71.80}}
& \aucg{81.45}{\textbf{82.89}} \\

MAGSAC++ \cite{DBLP:conf/cvpr/BarathNIM20}
& \aucg{32.39}{\textbf{32.69}}
& \aucg{60.24}{\textbf{61.03}}
& \aucg{71.34}{\textbf{71.97}}
& \aucg{82.46}{\textbf{82.86}} \\
\bottomrule
\end{tabular*}
\end{table}

\paragraph{Error Distribution Modeling.}
Tab.~\ref{tab:ablation_calibration} compares our Laplace mixture with alternative distributions and gating variants on MegaDepth-1500.
For scalar calibration diagnostics, we summarize the 2D residual with $e=\frac{1}{2}\|\mathbf r\|_1
=\frac{1}{2}(|r_x|+|r_y|)$ and report ECE$_{\rm err}$ between predicted $\mathbb{E}[e]$ and observed $e$, together with their Spearman rank correlation $\rho_s$.

The large gap between the fine-only model and all mixture variants in NLL and $\rho_s$ shows the importance of modeling coarse-assignment failures. The uncalibrated LMM gives the highest error-ranking correlation, but substantially worse NLL and ECE$_{\rm err}$, indicating mismatched uncertainty scales. Our calibration corrects this mismatch while largely preserving the ranking, achieving the best overall and heavy-tail likelihood with competitive scalar calibration. Removing or hardening the confidence gate weakens the likelihood. Fig.~\ref{fig:tail_behavior} further shows that the model assigns greater tail mass to harder datasets and weaker matchers, consistent with more frequent coarse-assignment failures.


\begin{table}[t]
\centering
\caption{\textbf{Calibration ablation on held-out MegaDepth.} We report mean per-axis NLL, ECE$_{\rm err}$, and Spearman correlation $\rho_s$ between predicted and observed correspondence error. Mixture modeling substantially improves error ranking over fine-only uncertainty, while calibration corrects the predicted error scale.}
\label{tab:ablation_calibration}
\vspace{-4pt}
\scriptsize
\renewcommand{\arraystretch}{1.12}
\begin{tabular*}{\linewidth}{@{\extracolsep{\fill}}lcccccc}
\toprule
\multirow{2}{*}{\textbf{Error model}} &
\multicolumn{4}{c}{\textbf{NLL} $\downarrow$} &
\multirow{2}{*}{\textbf{ECE$_{\rm err}$} $\downarrow$} &
\multirow{2}{*}{\textbf{$\boldsymbol{\rho_s}$} $\uparrow$} \\
\cmidrule(lr){2-5}
& \textbf{All} & \textbf{$<8$} & \textbf{$8$--$64$} & \textbf{$>64$} & & \\
\midrule
Fine only
    & 2.897
    & 2.808
    & \bestcell{4.373}
    & 11.749
    & 7.756
    & 0.240 \\

Gaussian mixture
    & \thirdcell{1.665}
    & \thirdcell{1.516}
    & 6.092
    & 12.158
    & 4.191
    & 0.348 \\

LMM, w/o calibration
    & 1.962
    & 1.874
    & \secondcell{4.762}
    & \secondcell{7.748}
    & 2.530
    & \bestcell{0.368} \\

LMM, w/o conf. gate
    & \secondcell{1.554}
    & \secondcell{1.440}
    & 5.724
    & \thirdcell{7.914}
    & \bestcell{1.575}
    & \thirdcell{0.355} \\

LMM, hard gate
    & 1.724
    & 1.552
    & 7.300
    & 13.100
    & \secondcell{1.579}
    & 0.347 \\

\textbf{Ours}
    & \bestcell{1.546}
    & \bestcell{1.437}
    & \thirdcell{5.532}
    & \bestcell{7.694}
    & \thirdcell{1.612}
    & \secondcell{0.360} \\
\bottomrule
\end{tabular*}
\vspace{-10pt}
\end{table}

\subsection{Robustness Analysis}

We follow the same calibration protocol as Sec.~\ref{subsec:settings}, but calibrate the model at \(960{\times}640\) to test robustness against both upscale and downscale. We then evaluate the learned parameters without retuning on the MegaDepth-1500 split across multiple input resolutions: $640{\times}480$, $480{\times}320$, and $1280{\times}960$. 

As shown in the top row of Fig.~\ref{fig:pdf_transfer}, the calibrated mixture tracks both the residual distribution and observed error across resolutions, whereas fine-only uncertainty remains bounded by the local refinement window and becomes overconfident for large errors.

The reliability diagram in the bottom row of Fig.~\ref{fig:pdf_transfer} shows that our calibrated mixture provides a substantially better estimate of error than the fine-only baseline, correctly assigning high uncertainty to severe mismatches. Conversely, the fine-only model is strictly bounded by its refinement window, unable to predict error greater than roughly 10px. It therefore outputs severely overconfident, low uncertainties for massive spatial outliers.

\subsection{Computational Analysis}
\label{subsec:runtime}

\begin{figure}[t]
  \centering
    \caption{\textbf{Runtime overhead.} The total latency added by uncertainty extraction and CoRe. The method introduces minimal overhead, acting as a lightweight post-processing step.}
    \vspace{-10pt}
  \includegraphics[width=0.9\linewidth]{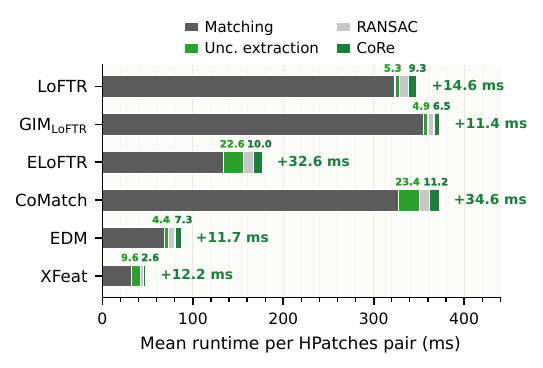}
  \label{fig:runtime_bar}
\vspace{-25pt}
\end{figure}

CoRe introduces minimal runtime overhead by leveraging existing intermediate outputs from the pretrained matcher. This process adds very little additional cost.

Runtimes were evaluated on an NVIDIA RTX A5000 GPU, averaged over 100 image pairs from the HPatches dataset \cite{DBLP:conf/cvpr/BalntasLVM17}. As shown in Fig.~\ref{fig:runtime_bar}, uncertainty extraction and CoRe add only
$11.4$--$34.6$ms across matchers despite our unoptimized NumPy
implementation.
Fitting the overall error model is also highly efficient, requiring only 9 learnable calibration parameters.
\section{Conclusion}
In this work, we demonstrate that semi-dense matching uncertainty is not just local confidence. We present a novel overall uncertainty estimation method that incorporates both coarse-level matching failures and refinement errors. Extensive experiments show our approach achieves significant performance improvements. These results highlight the practical value of overall uncertainty modeling as a simple yet highly effective step toward reliable image matching. 
\clearpage
\setcounter{section}{0}

{
    \small
    \bibliographystyle{ieeenat_fullname}
    \bibliography{main}
}
\clearpage
\appendix
\setcounter{section}{0}

\twocolumn[{
\begin{center}
    {\Large \bfseries Semi-Dense Matching Uncertainty Is Not Just Local Confidence\par}
    \vspace{0.7em}
    {\large Supplementary Material\par}
\end{center}
\vspace{1.5em}
}]

This supplementary provides additional analyses, qualitative results, and implementation details for our method. Unless otherwise stated, all additional experiments use EfficientLoFTR \cite{wang2024eloftr} as the base matcher and MegaDepth \cite{MDLi18} as the calibration set.
\section{Additional Implementation Details}

\paragraph{Axis-wise mixture implementation.}
\label{sec:axis-implementation}
In practice, we use an axis-factorized implementation with separate one-dimensional Laplace mixtures for the horizontal and vertical residuals. The resulting two-dimensional likelihood is
\begin{equation}
\begin{aligned}
\hat p_i(\mathbf r_i)
=
\prod_{d\in\{x,y\}} \hat p_{i,d}(r_{i,d})
\end{aligned}
\end{equation}
Our parameter set $\Theta=\{a_x,a_y,b_x,b_y,k_s,k_m,t_x,t_y,t_m\}$ defines the axis-wise likelihood: 
\begin{equation}
\begin{aligned}
\hat p_{i,d}(r_{i,d})
=
(1-\alpha_{i,d})
\operatorname{Lap}(r_{i,d}\mid 0,s_{i,d}^{f}) \\
+ \alpha_{i,d} \operatorname{Lap}(r_{i,d}\mid 0,s_{i,d}^{c}),
\end{aligned}
\end{equation}
where
\[
\begin{cases}
s_{i,d}^{f}
&=
\sqrt{b_d}\,\tilde{s}_{i,d}^{f},\\
s_{i,d}^{c}
&=
\sqrt{a_d}\,\tilde{s}_{i,d}^{c},\\
\alpha_{i,d}
&=
\sigma\!\left[
k_s\!\left(\bar{s}_{i,d}^{c}-t_d\right)
+
k_m\!\left(-\log m_i-t_m\right)
\right],\\
\bar{s}_{i,d}^{c}
&=
100\,\frac{\tilde{s}_{i,d}^{c}}{D_d},
\qquad D_x=W,\quad D_y=H
\end{cases}
\]
Here, $\tilde{s}_{i,d}^{f}$ and $\tilde{s}_{i,d}^{c}$ denote the raw fine- and coarse-level standard deviations, respectively, $\bar{s}_{i,d}^{c}$ denotes the image-size-normalized coarse scale, and $m_i$ denotes the matching confidence.

\paragraph{Uncertainty extraction.}
We extract the coarse uncertainty cues consistently across all matchers. For each selected coarse match, the coarse scale $\tilde s_i^c$ is computed as the standard deviation of the row-wise softmax distribution used to form the dual-softmax score matrix, while the corresponding selected entry provides the matching confidence \cite{sun2021loftr, wang2024eloftr}. For XFeat \cite{DBLP:conf/cvpr/PotjeC0MN24}, whose similarity matrix is computed from unnormalized features, we additionally normalize the features and apply the softmax explicitly. 
This accounts for its slightly higher extraction overhead reported in Sec.~\ref{subsec:runtime}.

Fine uncertainty scale $\tilde s_i^f$ is extracted from each matcher's native refinement distribution. For LoFTR \cite{sun2021loftr}, GIM-LoFTR \cite{xuelun2024gim}, and XFeat \cite{DBLP:conf/cvpr/PotjeC0MN24}, we compute the standard deviation of the local $8{\times}8$ heatmap. For EDM \cite{DBLP:conf/iccv/LiRP25}, it is obtained from the axis-wise refinement distributions. EfficientLoFTR~\cite{wang2024eloftr} and CoMatch \cite{DBLP:conf/iccv/LiLTZM25} employ two-stage correlation refinement; we therefore extract and combine the uncertainty from both refinement stages, resulting in their higher extraction cost.

\paragraph{Calibration optimization.}
We jointly optimize the calibration parameters by minimizing the mixture negative log-likelihood on the held-out calibration set. We use bounded, derivative-free Powell optimization \cite{powell1964efficient} with a maximum of $120$ iterations and a function-value tolerance of $10^{-7}$. 

\section{Additional Analysis}

\subsection{Sensitivity to Initial Geometry}
\label{subsec:initial_geometry}

\begin{figure}[t]
  \centering
    \caption{\textbf{Sensitivity to initial geometry across matchers.} Top: median relative ACE reduction after CoRe across ranges of initial ACE $e_0$. Bottom: fraction of pairs improved by CoRe. The marker area reflects the number of valid pairs in each bin.}
    \vspace{-4pt}
    \includegraphics[width=0.97\linewidth]{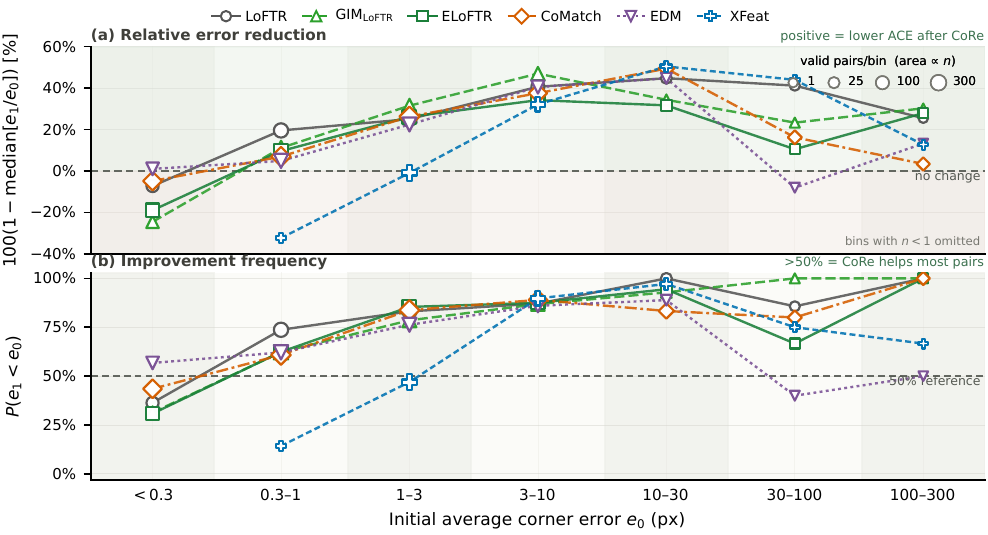}
  \label{fig:initial_geometry_sensitivity}
  \vspace{-10pt}
\end{figure}

During inference, CoRe estimates the coarse-success posterior from residuals induced by the initial geometry $\theta_0$ returned by the robust estimator, rather than from the ground-truth matching errors available during calibration. Specifically,
\begin{equation}
\mathbf r_i^{(0)}
=
\hat{\mathbf y}_i-\Pi_{\theta_0}(\mathbf x_i)
=
\underbrace{\hat{\mathbf y}_i-\mathbf y_i}_{\mathbf r_i}
+
\underbrace{\mathbf y_i-\Pi_{\theta_0}(\mathbf x_i)}_{\text{initial error}},
\end{equation}
When $\theta_0$ is accurate, $\mathbf r_i^{(0)}$ closely approximates the true matching error $\mathbf r_i$. As the initial estimate deteriorates, the additional geometry error may distort the inferred posterior. We therefore use the average corner error of the initial homography $e_0$ as an indicator of initialization quality and compare it with the error after CoRe $e_1$.

Table~\ref{tab:initial_geometry_quality} shows that CoRe improves geometry estimation over a wide range of initial errors. The strongest and most consistent gains occur for $e_0$ between $1$ and $30$ px, where CoRe improves $85.4$--$94.4\%$ of image pairs and produces increasingly larger median error reductions as the initial estimate deteriorates. Positive median changes in the number of geometric inliers further indicate that the refit recovers a more consistent correspondence set.

Figure~\ref{fig:initial_geometry_sensitivity} shows the same trend across matchers and mirrors the findings in the main paper. CoRe is most beneficial for moderately difficult pairs, where coarse-assignment failures remain frequent enough to affect geometry estimation. For easy pairs with very small initial error, most matches are already correct and explicitly modeling coarse uncertainty provides little additional benefit. The behaviour becomes less stable for very poor initial estimates above $30$ px, where substantially fewer valid pairs are available.

\begin{table}[t]
\centering
\caption{\textbf{Sensitivity to initial geometry for EfficientLoFTR on HPatches.} Pairs are grouped by the ACE of the initial homography $e_0$. We report the median reduction $\Delta e=e_0-e_1$, the fraction of improved pairs, and the median pairwise change in geometric inliers after CoRe.}
\label{tab:initial_geometry_quality}
\vspace{-4pt}
\scriptsize
\setlength{\tabcolsep}{1.5pt}
\renewcommand{\arraystretch}{1.08}
\providecommand{\inlg}[2]{#1{\scriptsize$\!\rightarrow\!$}#2}
\begin{tabular*}{\linewidth}{@{\extracolsep{\fill}}lcccc}
\toprule
\textbf{Initial ACE $e_0$} &
\textbf{\# pairs} &
\textbf{Median $\Delta e$ (px)} $\uparrow$ &
\textbf{Improved rate} $\uparrow$ &
\textbf{Median $\Delta$ inliers} $\uparrow$ \\
\midrule
$<1$ px
& 261 & $-0.00$ & 47.9\% & $+5$ \\

$1$-$3$ px
& 144 & $+0.42$ & 85.4\% & $+78$ \\

$3$-$10$ px
& 96 & $+1.65$ & 87.5\% & $+88.5$ \\

$10$-$30$ px
& 18 & $+4.26$ & 94.4\% & $+39$ \\

$30$-$100$ px
& 6 & $+6.61$ & 66.7\% & $+33$ \\

$100$-$300$ px
& 3 & $+57.7$ & 100.0\% & $+28$ \\

$\geq300$ px                
& 12 & $+735.6$ & 58.3\% & $0$ \\
\bottomrule
\end{tabular*}
\end{table}

\subsection{Coarse-Success Posterior Quality}
\label{subsec:posterior_quality}

We further examine whether the inferred coarse-success posterior
$\Pr(\hat{\mathbf y}_i^c=\mathbf y_i^{c,*}\mid\mathbf r_i^{(0)})$
captures the actual correctness of the predicted coarse assignment. We group matches by their posterior probability and compare each bin with its empirical coarse-assignment success rate on HPatches \cite{DBLP:conf/cvpr/BalntasLVM17}.

\begin{figure}[t]
  \centering
    \includegraphics[width=0.97\linewidth]{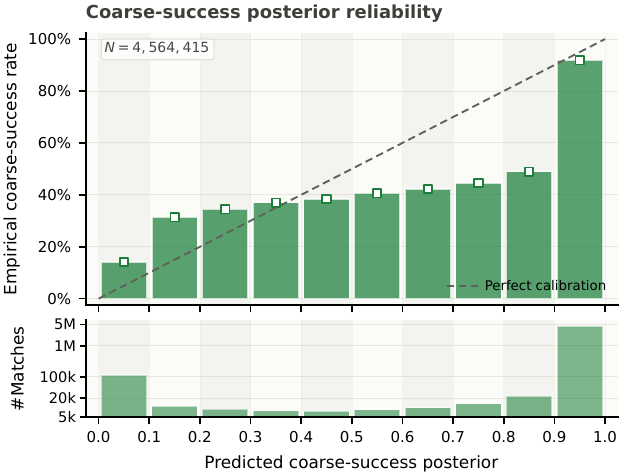}
    \caption{\textbf{Coarse-success posterior quality on HPatches.}
    Matches are grouped by predicted posterior probability and compared with their empirical coarse-assignment success rate. Higher posterior values consistently correspond to more accurate coarse assignments.}
  \label{fig:posterior_reliability}
\end{figure}

\begin{figure}[t]
  \centering
    \includegraphics[width=1.0\linewidth]{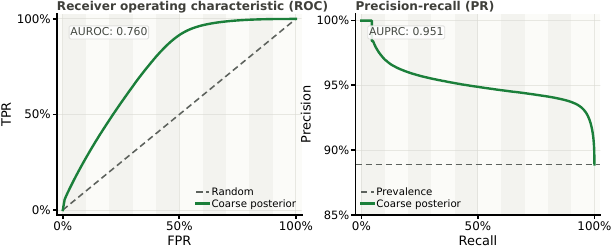}
    \caption{\textbf{Coarse-success discrimination on HPatches.}
    ROC and precision--recall curves of the predicted posterior, achieving AUROC $=0.760$ and AUPRC $=0.951$.}
  \label{fig:posterior_roc_pr}
\end{figure}

Figure~\ref{fig:posterior_reliability} shows a clear monotonic relationship between predicted posterior and empirical success frequency. Although the probabilities are not perfectly aligned, this is under zero-shot transfer from MegaDepth~\cite{MDLi18} calibration to HPatches. Figure~\ref{fig:posterior_roc_pr} further reports AUROC $=0.760$ and AUPRC $=0.951$, providing additional quantitative evidence that the posterior meaningfully discriminates coarse successes from failures and supports its use as the correspondence weight in CoRe.

\subsection{Calibration Data Quantity}
A primary advantage of our post-hoc calibration framework is that it avoids the massive computational cost associated with retraining the base matching networks. To validate the efficiency and stability of our method, we investigate the impact of the calibration set size on the learned mixture parameters.

\begin{figure}[t]
  \centering
    \includegraphics[width=1.\linewidth]{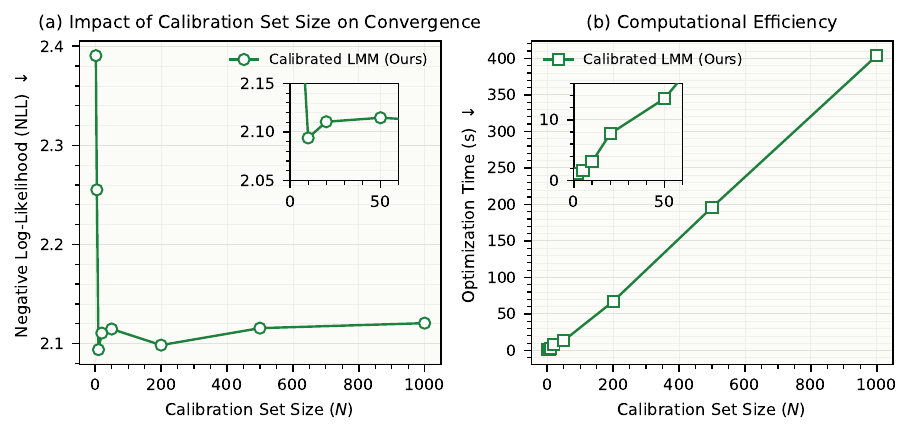}
    \caption{\textbf{Impact of calibration data quantity on convergence and computational efficiency.} (a) The calibration objective function (Negative Log-Likelihood) over varying numbers of calibration pairs. (b) Optimization time, which scales almost linearly with the number of samples but remains under a few seconds in the practical low-data regime.}
  \label{fig:data_impact_convergence}
\end{figure}

We randomly sample subsets of varying sizes, denoted as $N \in \{2, 5, 10, 20, 50, 200, 500, 1000\}$, from the held-out split of MegaDepth \cite{MDLi18}. For each calibration subset size, we independently run our calibration algorithm to estimate the calibration parameters \(\Theta=\{\mathbf a,\mathbf b,\mathbf w,\beta\}\) for the Laplace mixture model.
As illustrated in Fig. \ref{fig:data_impact_convergence}, the optimization process exhibits rapid initial convergence, with the NLL dropping sharply between 2 and 50 pairs as the algorithm quickly captures the core spatial error modalities. Beyond 50 pairs, the NLL curve achieves stabilization and convergence. Notably, because even a single pair of images typically yields approximately 2,000 correspondence residuals, our calibration algorithm is effectively fed with an abundant set of correspondence constraints even at the smallest data scale.

Furthermore, because $\Theta$ contains only 9 learnable parameters, calibration is highly efficient, requiring little calibration data and optimization time. Using Powell's method as implemented in SciPy~\cite{2020SciPy-NMeth}, optimization on 200 image pairs takes 67.2 seconds, with runtime scaling approximately linearly with calibration-set size.

\begin{figure}[t]
  \centering
    \includegraphics[width=1.\linewidth]{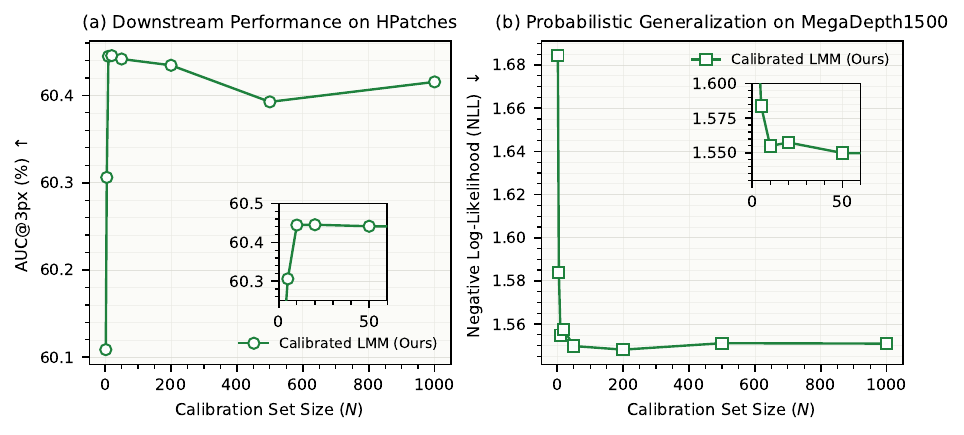}
    \caption{\textbf{Impact of calibration set size on downstream performance and generalization.} We evaluate the HPatches AUC@3px (a) and mean per-axis NLL on MegaDepth1500 (b) as functions of the number of calibration pairs ($N$). Our calibration algorithm demonstrates exceptional sample efficiency, achieving convergence earlier with only 10 pairs.}
  \label{fig:data_impact_validation}
\end{figure}

Finally, we analyze the impact of calibration data quantity on final downstream performance, specifically using AUC@3px on HPatches \cite{DBLP:conf/cvpr/BalntasLVM17} and the mean per-axis NLL on MegaDepth1500 \cite{MDLi18}, as shown in Fig. \ref{fig:data_impact_validation}. Our results consistently demonstrate that the calibration algorithm reliably converges within a practical range of 10 to 200 image pairs. Notably, even when the calibration set size is increased significantly beyond this range, the performance metrics remain highly stable. This empirical evidence confirms the numerical robustness of our formulation, maintaining consistent performance regardless of the calibration data size.

\subsection{Relative Pose Estimation Study}
\label{sec:supp_relative_pose}

We did not include relative pose estimation in the main evaluation because its geometry differs fundamentally from homography estimation and PnP. In the latter tasks, an estimated geometric model provides a point-wise projection $\Pi_{\theta}(\mathbf{x}_i)$, allowing CoRe to compute a 2D residual
\begin{equation}
    \mathbf{r}_i =
    \hat{\mathbf{y}}_i -
    \Pi_{\theta}(\mathbf{x}_i),
\end{equation}
which directly approximates the correspondence error when the initial geometry is sufficiently accurate. In contrast, an essential matrix maps each point only to an epipolar line rather than to a unique point in the second image. Therefore, the full 2D correspondence residual required by CoRe is not directly available.

We additionally consider a \emph{depth-assisted} setting, where the ground-truth source depth $z_i$ and translation scale $\|t_{\rm gt}\|$ are given, while the rotation $\hat R$ and translation direction $\hat t$ are predicted from the initial essential matrix. This gives the point-wise projection
\begin{equation}
    \Pi_{\theta}(\mathbf{x}_i)
    =
    \pi\!\left(
    \hat R\left(z_i\tilde{\mathbf{x}}_i\right)
    +
    \|t_{\rm gt}\|\hat t
    \right),
\end{equation}
where $\tilde{\mathbf{x}}_i$ denotes the homogeneous image coordinate under the canonical camera $K=I$, and $\pi(\cdot)$ denotes perspective division. From this, CoRe can compute the full 2D reprojection residual. This setting is used only to study the effect of having a valid point-wise reprojection, rather than as a practical relative-pose pipeline.

\begin{table}[t]
\centering
\caption{\textbf{Relative pose estimation ablation on MegaDepth-1500.}
Pose AUC under standard and depth-assisted settings.}
\label{tab:relative_pose}
\vspace{-4pt}
\scriptsize
\renewcommand{\arraystretch}{1.12}
\begin{tabularx}{\linewidth}{
    @{}
    l
    c
    *{4}{>{\centering\arraybackslash}X}
    @{}
}
\toprule
\multirow{2}{*}{\textbf{Method}} &
\multirow{2}{*}{\textbf{GT depth}} &
\multicolumn{4}{c}{\textbf{Pose Estimation AUC} $\uparrow$} \\
\cmidrule(lr){3-6}
& & \textbf{@$1^\circ$} & \textbf{@$3^\circ$}
  & \textbf{@$5^\circ$} & \textbf{@$10^\circ$} \\
\midrule

Base
& \multirow{2}{*}{No}
& 8.06
& \bestcell{28.52}
& \bestcell{41.81}
& \bestcell{60.42} \\

w/ epipolar-based CoRe
&
& \bestcell{8.91}
& 21.32
& 28.51
& 38.98 \\

\midrule

w/ depth-assisted Huber
& \multirow{2}{*}{Yes}
& 12.14
& 28.87
& 38.58
& 52.48 \\

\textbf{w/ depth-assisted CoRe}
&
& \bestcell{26.42}
& \bestcell{45.94}
& \bestcell{54.24}
& \bestcell{64.75} \\

\bottomrule
\end{tabularx}
\vspace{-8pt}
\end{table}

Table~\ref{tab:relative_pose} supports the above interpretation. Using the epipolar distance as a surrogate for the correspondence residual substantially degrades performance at most thresholds. This is expected, as the point-to-line distance captures only the component of the correspondence error normal to the epipolar line while completely discarding displacement along the line. In contrast, when a valid point-wise reprojection is available in the depth-assisted setting, CoRe becomes effective, with consistent gains against one-step Huber~\cite{Huber1992} across all evaluated thresholds. These results indicate that CoRe is most suitable for problems where the geometric model provides a point-wise reprojection from which the full 2D correspondence residual can be computed.

\subsection{Comparison with Native Matcher Uncertainty}

\begin{table}[t]
\centering
\vspace{-4pt}
\scriptsize
\renewcommand{\arraystretch}{1.12}
\begin{tabular*}{\linewidth}{@{\extracolsep{\fill}}llcccc}
\toprule
\multirow{2}{*}{\textbf{Method}} &
\multirow{2}{*}{\textbf{Metric}} &
\multicolumn{4}{c}{\textbf{GT error threshold (px)}} \\
\cmidrule(lr){3-6}
& & $\boldsymbol{\tau} =8$ &  $\boldsymbol{\tau} =32$ &
 $\boldsymbol{\tau} =64$ &  $\boldsymbol{\tau} =128$ \\
\midrule

\multirow{2}{*}{PDC-Net+}
& AUROC $\uparrow$
    & \secondcell{89.73}
    & 93.54
    & 94.19
    & \secondcell{95.13} \\
&GT Error Rate
    & 9.58\%
    & 5.52\%
    & 4.21\%
    & 2.77\% \\

\addlinespace[2pt]
\multirow{2}{*}{RoMaV2}
& AUROC $\uparrow$
    & \bestcell{95.82}
    & \bestcell{96.60}
    & \secondcell{96.49}
    & 96.24 \\
&GT Error Rate
    & 8.52\%
    & 6.18\%
    & 5.43\%
    & 4.50\% \\

\addlinespace[2pt]
\multirow{2}{*}{\textbf{Ours (ELoFTR)}}
& AUROC $\uparrow$
    & 88.64
    & \secondcell{95.73}
    & \bestcell{96.87}
    & \bestcell{97.88} \\
&GT Error Rate
    & 2.70\%
    & 0.51\%
    & 0.32\%
    & 0.20\% \\

\bottomrule
\end{tabular*}
\caption{\textbf{Comparison with native learned uncertainty on HPatches.}
AUROC measures how well uncertainty identifies matches whose GT error exceeds
each threshold; Error reports their fraction (\%).}
\label{tab:native_uncertainty}
\vspace{-10pt}
\end{table}

We further compare our calibrated uncertainty with matchers that directly learn global pixel-space predictive uncertainty, namely PDC-Net+~\cite{truong2023pdc} and RoMaV2~\cite{edstedt2025romav2}.
Because these are dense matchers operating under a different matching paradigm, they produce different correspondence sets and error distributions. We therefore focus on how well uncertainty identifies each matcher's own correspondences. For each threshold $\tau$, we report AUROC for predicting whether the GT correspondence error exceeds $\tau$, together with the corresponding fraction of errors.

\begin{figure}[t]
  \centering
  \includegraphics[width=0.97\linewidth]{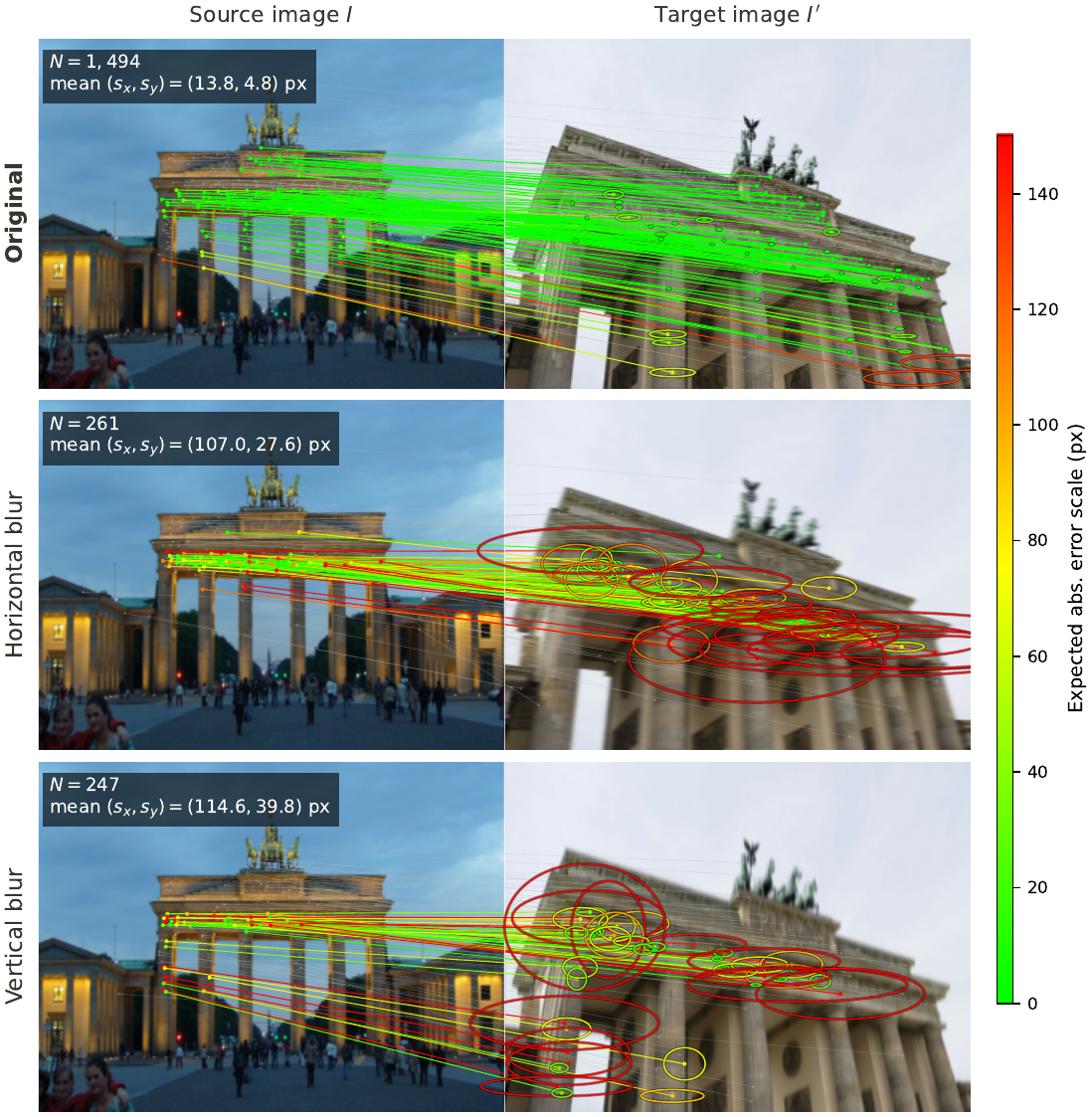}
  \caption{\textbf{Uncertainty under controlled directional blur.}
  Predicted spatial error scales are visualized as ellipses around the matches. Horizontal and vertical Gaussian blur increase the estimated uncertainty primarily along the corresponding blur direction.}
  \label{fig:blur_uncertainty}
\end{figure}

All methods are evaluated on HPatches~\cite{DBLP:conf/cvpr/BalntasLVM17} at $1184\times896$ resolution. For the dense matchers, we randomly sample $8.5$K correspondences per pair to obtain a comparable evaluation size to ELoFTR. As shown in Tab.~\ref{tab:native_uncertainty}, RoMaV2 is strongest at
moderate errors, while our lightweight post-hoc uncertainty becomes competitive for larger failures and achieves the highest AUROC at $64$ and $128$px. These results show that, despite requiring no matcher retraining, carefully calibrated post-hoc uncertainty can remain competitive with jointly learned uncertainty estimates.

\section{Additional Qualitative Results}
\label{sec:qualitative}

\subsection{Uncertainty Changes under Controlled Blur}

Figure~\ref{fig:blur_uncertainty} visualizes the absolute spatial error scales predicted by the calibrated mixture as ellipses around the matches. We apply horizontal and vertical Gaussian blur to the target image. The predicted ellipses expand primarily along the corresponding blur direction, showing that the estimated uncertainty responds to directional image degradation as expected.

\subsection{More CoRe Refinement Examples}
\begin{figure*}[t]
      \centering
    \includegraphics[width=0.97\linewidth]{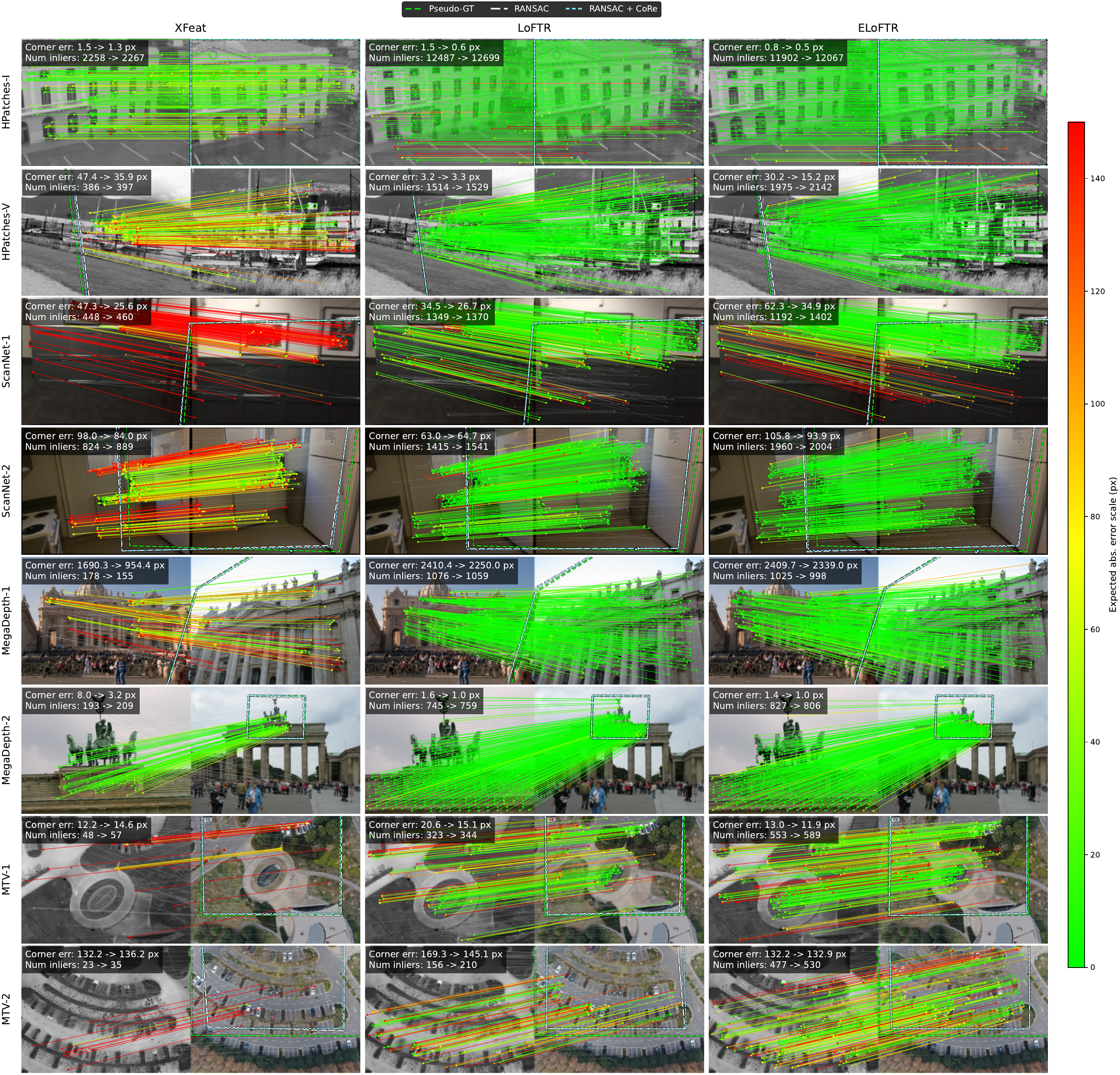}
    \caption{\textbf{Additional qualitative CoRe refinement results.}
    We compare the initial and CoRe-refined homographies across four datasets and three representative matchers. CoRe improves alignment in most examples, while some challenging MTV cases illustrate representative failure modes.}
  \label{fig:big_qualiative}
\end{figure*}

A full page of qualitative CoRe results across four datasets and three representative matchers is shown in Fig.~\ref{fig:big_qualiative}. CoRe improves homography estimation in most cases, although several MTV \cite{DBLP:journals/remotesensing/LiuLYCZPZ23} examples remain challenging and are included as representative failure cases.

\end{document}